# A Novel Remote Sensing Approach to Recognize and Monitor Red Palm Weevil in Date Palm Trees (manuscript)

Yashu Kang[1], Chunlei Chen[1], Fujian Cheng[1], Jianyong Zhang[1]

[1] STAR VISION

March 20, 2022

**Abstract**

The spread of the Red Pal Weevil (RPW) has become an existential threat for palm trees around the world. In the Middle East, RPW is causing wide-spread damage to date palm Phoenix dactylifera L., having both agricultural impacts on the palm production and environmental impacts. Early detection of RPW is very challenging, especially at large scale. This research proposes a novel remote sensing approach to recognize and monitor red palm weevil in date palm trees, using a combination of vegetation indices, object detection and semantic segmentation techniques. The study area consists of date palm trees with three classes, including healthy palms, smallish palms and severely infected palms. This proposed method achieved a promising 0.947 F1 score on test data set. This work paves the way for deploying artificial intelligence approaches to monitor RPW in large-scale as well as provide guidance for practitioners.

**Key words:** red palm weevil, remote sensing, satellite imagery

## 1. Introduction

The Red Palm Weevil (RPW), also known as Rhynchophorus Ferrugineus and Rhynchophorus Vulneratus) is a type of beetle that attacks different species of palm trees and has become an existential threat for palm trees around the world. In the Middle East, RPW is causing wide-spread damage to date palm Phoenix dactylifera L., having both agricultural impacts on the palm production and environmental impacts.

Often targeting young trees, the mechanism of infection involves the beetles laying eggs inside with the eventual larvae feeding on the palm's tissue, thus creating tunnels inside the tree trunk that weaken its structure, finally causing extensive damage that results in decline and even breakage of the tree (Kagan, et al., 2021). In fact, the RPW has become a worldwide concern despite originating from tropical areas in Asia (Gerson and Applebaum, 2020). According to the EPPO (European and Mediterranean Plant Protection Organization) datasheet, the RPW has spread to 85 different countries and regions worldwide (Eppo, 2021). Moreover, in many countries, palm trees are considered decorative trees planted in residential neighborhoods, and the RPW threat may lead to trees breaking down and hence a higher likelihood of endangering human lives (Kagan, et al., 2021).

Early detection of RPW infestation is vital for a more effective control of the pest damage. However, the entire life cycle of RPW larvae is hard to observe from outside. To date, the lifecycle of RPW and relevant biological parameters varies in different regions. Some established findings are summarized in Table 1 below.

Table 1. The development period of RPW.

| Source | Development period (days) | | | |
|---|---|---|---|---|
| | Egg | Larvae | Pupae | Total |
| El-Shafie et al. | 3-5 | 33-46 | 20-36 | - |
| Ju et al. | 3-4 | 30-67 | 23-36 | - |
| Salama et al. | - | 69-128 | 16-29 | - |
| Faghih et al. | 1-6 | 41-78 | 15-27 | - |
| Kalshoven et al. | - | 44-210 | | 105-210 |
| Marzukhi et al. | 2-5 | 25-105 | 14-21 | |
| FAO | 2-5 | 35-129 | - | - |

Conventional methods might be costly in some cases (e.g., acoustic detection). A lot of in-field detections were achieved through the observation of symptoms. However, as symptoms only show in late stages, the detection and prediction of RPW infestations in early stages remains a significant challenge. This study proposes a novel remote sensing approach to recognize and monitor RPW infestation that combinates artificial intelligence algorithm and spectral analysis. The rest of the article is organized as follows: section 2 presents a review of related studies. Section 3 describes the used methodology while section 4 demonstrates the experimental results. Lastly, the findings are concluded and discussed in Section 5.

**2. Literature Review**

Early detection of palm tree infestation is vital and may save trees from irreversible damage. At those earlier stages, control methods could be used, most commonly by the application of insecticides (FAO, 2020). Since today, there are various techniques applied to detect RPW, such as visual inspection, chemical detection, acoustics monitoring, remote sensing (multispectral and thermal imaging), etc. Visual inspection depends on the stage of infestation because the appearances only become clearly visible until it gets severe or the palm tree starts to break. In chemical detection, a trained dog or electronic nose can discover early stage RPW with high accuracy, with the disadvantage of training cost and specific types of dogs. In terms of acoustic monitoring, sensors can be used to examine the sounds from pest activity as the weevils are feeding on the tissues of the trunk (Koubaa, 2019).

Apart from the conventional approaches mentioned above, remote sensing attempts to obtain and analyze data without physically interacting with the object. Through remote sensing one aims to get spectrum information of each pixel in an image. By mounting sensors onto a satellite or UAV and measuring the radiation emitted from the tree surface, it is non-invasive, affordable and tailored for monitoring at large scale. Several spectral indices such as vegetation, biomass and chlorophyll could be investigated to help explore bio-physiological variations. Reflectance values collected from the crown trees are heavily associated with the pigment composition in the leaves including chlorophylls, carotenoids, water content, and flavonoids. Therefore, it is believed that different stages of RPW can be differentiated using a combination of empirical data and spectral information. For instance, the Sentinel 2A satellite provides easily accessible data with blue, green, red (RGB) and near-infrared (NIR) bands at 10-meter spatial resolution. The visible RGB bands show the effects of pigmentation constituents such as chlorophyll and carotenoids; whereas the red-edge and NIR bands express the biomass status, while the eight SWIR bands illustrate the water content change progressively from the healthy to the dead palm trees (Bannari et al., 2018).

Among various types of vegetation indices, NDVI (Normalized Differenced Vegetation Index) is the most frequently used in existing studies. However, in high vegetation covered areas, NDVI may be saturated, in addition to its non-linear relationship with bio-physiological variations.

By combining WorldView-3 multispectral imagery and in-field samples, Abdou et al. explored statistical analysis such as linear and second order polynomial regression to monitor different RPW infection levels: dead, severely attacked, attacked-untreated, attacked-treated and healthy (Abdou et al., 2017). A total of eleven indices of water stress were extracted from eighty samples, from which regression models were built to estimate the relationship between water stress and water content dynamic range. They concluded most water stress indices were important and achieved R-squared values higher than 0.95, except Palm Tree Water Stress Index (PTWSI).

Bannari et al. also adopted WorldView-3 satellite data along with bidirectional reflectance through spectroradiometer (Bannari et al., 2018). Water content information collected from one hundred in-field samples was quantified, calibrated and compared with lab results for statistical analysis. The findings indicated PTWSI-4 (Palm Tree Water Stress Indices) and SRWI (Simple Ratio Water Index) indices could help discriminate among different levels of stress attack with confidence level at 95%. Based on their results, remote sensing is a promising alternative technology for RPW detection based on water stress indices.

Carter suggested that crop canopy water content was significantly correlated with NIR (720-1000 nm) and SWIR (1000-2500 nm) reflectance (Carter, 1991). After removal of background signals related to the canopy structures, spectral derivation showed that the estimation of plant water content is improved at wavelengths of intermediate water absorption (1450 nm), while wavelengths of near-total water absorption (1940 and 2500 nm) showed little to no sensitivity to variations in liquid water content (Bannari et al., 2018).

Examining characteristics of the reflectance spectra of palm trees with known infestation levels of healthy, moderately infected and severely infected, Yones et al. found that specific waveband zones were effective to differentiate these levels (Yones et al., 2014). For instance, the optical spectral ranges of [514-664nm] and [684-1344 nm] were sufficient to identify healthy trees; [529-589 nm], [693-695 nm] and [1333-1335 nm] could be used to identify moderately infected ones; using [720-724 nm] alone was able to recognize severely infected date palm trees.

Due to breakthroughs in various fields and recent progress in very-high-resolution (VHR) aerial images, more and more studies started to adopt computer vision techniques in aiding recognition of palm trees. Kagan et al. analyzed more than 100,000 aerial and street view images to develop a deep learning approach to detect and map palm trees with RPW infestation (Kagan et al., 2021). The change of water conductivity in severely infected samples resulted in drying the outer leaves and fruit bunches as well as toppling of the trunk. However, their study depended on vision-based information to differentiate infected samples, which may not be suitable for early detection.

Kurdi et al. evaluated performances of 10 classification approaches in using plant-size and temperature measurements collected from individual trees to predict RPW infestation in its early stages (Kurdi et al., 2021). Data was collected from date palm trees located at Kharj area of Riyadh, Saudi Arabia. They concluded that temperature and circumference were the most important features in predicting RPW infestation.

## 3. Methodology and Study Area

Figure 1 shows the study areas of this research, which are located in Riyadh, Saudi Arabia. Images acquired from WorldView-3 provided RGB bands with 0.31 m spatial resolution, NIR with 1.24 m spatial resolution, etc. In this research, palm tree samples are manually annotated with three classes: healthy palm, smallish palm and dead palm. As is shown in Figure. 2, healthy palms denote the palm that is normally cultivated. Smallish palms mean the diameter of the palm crown is relatively smaller than healthy and mature date palm trees, possibly under a seeding stage or under an abnormal growing stage. Dead palms no longer have vitality and usually have gray tree crowns or have broken down.

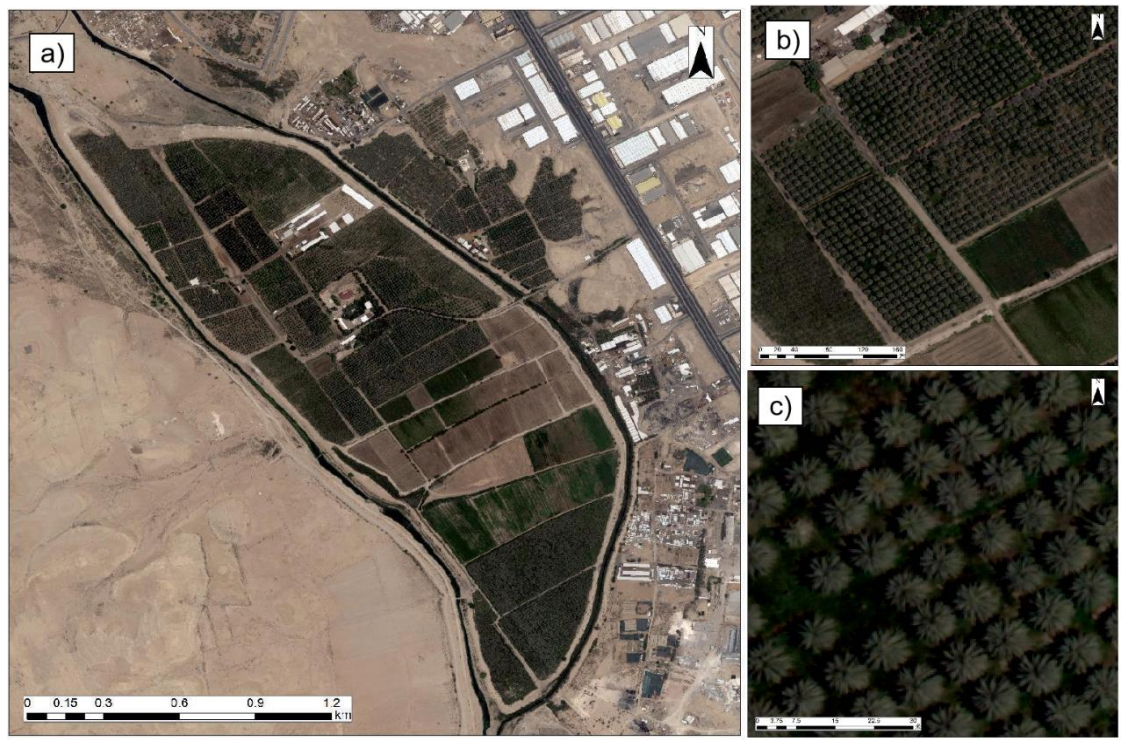

Figure 2 Study area

The whole dataset, consisting of 4,682 palm instances, was split into training data and test data, into 80/20 proportion. A machine learning model was trained and validated using these two data sets, respectively. The proposed method is a combination of object detection and semantic segmentation algorithms. The workflow is as follows:

1. Extract vegetation coverage areas using vegetation indices like NDVI
2. Divide original images into smaller instances with 1024x1024 pixels
3. Augment the training data using cropping, Gaussian or Gamma noises, flips, scaling, etc.
4. Use semantic segmentation algorithm to obtain the "probability map" of data palm trees
5. Apply end-to-end object detection algorithms to tree crown recognition with multiclass labels

The semantic segmentation model inherits a general U-Net architecture, as shown in the figure below, while the object detection part used a one-stage detector YOLOv5.

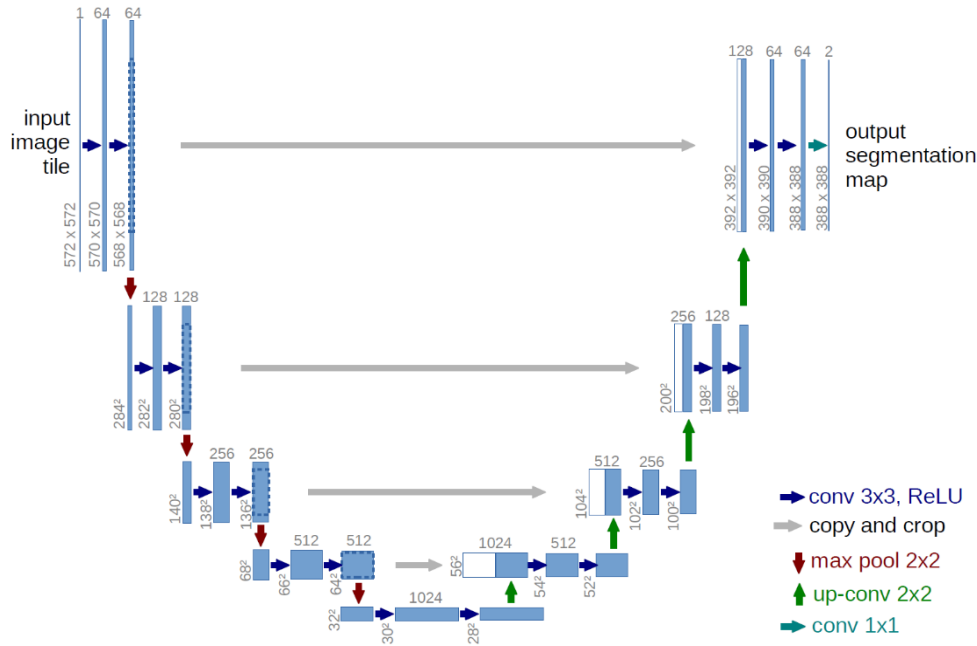

Figure 3 U-Net Architecture

Classification performance is assessed through a confusion matrix. A confusion matrix is an N x N matrix, where N is the number of target classes. The matrix compares the actual target values with those predicted by the machine learning model, as shown in Figure 4 below.

Figure 4 Confusion matrix

The basic terminology is as follows:

1) True Positives (TP): when the actual value is Positive and predicted is also Positive.
2) True negatives (TN): when the actual value is Negative and prediction is also Negative.
3) False positives (FP): When the actual is Negative but prediction is Positive. Also known as the Type I error
4) False negatives (FN): When the actual is Positive but the prediction is Negative. Also known as the Type II error.

In addition, calculations of classification metrics such as accuracy, precision, recall and F1 score are as follows:

$$Accuracy = \frac{TP + TN}{TP + TN + FP + FN}$$

$$Precision = \frac{TP}{TP + FP}$$

$$Recall = \frac{TP}{TP + FN}$$

$$F1 = 2 * \frac{Precision * Recall}{Precision + Recall}$$

Accuracy simply measures how often the classifier makes the correct prediction. It's the ratio between the number of correct predictions and the total number of predictions. Precision measures correctness that is achieved in true prediction, i.e., how many predictions are actually positive out of all the total positive predicted. Recall is a measure of actual observations which are predicted correctly, i.e., how many observations of positive class are actually predicted as positive. It is also known as Sensitivity. Recall is a valid choice of evaluation metric when we want to capture as many positives as possible. The F1 score is a number between 0 and 1 and is the harmonic mean of precision and recall.

Note that for the purpose of feature engineering, several vegetation indices such as NDVI and gNDVI were created using empirical knowledge. For instance, the calculation of gNDVI in this research is as follows:

$$gNDVI = \frac{NIR - G}{NIR + G}$$

As reported by Gitelson et al., gNDVI is functionally associated with only the chlorophyll content, without being sensitive to other pigments like carotenoid.

To determine the key features that contribute to the classification of a palm tree as infested or healthy, Pearson's correlation coefficient between the features and response variable were computed. Features with a value of zero for any of the computed measures were removed.

4. Analysis Results

Based on the vegetation indices tested in this research, the results shown in Figure 5 demonstrated potential for correspondence with field observation. Depending on infestation stages of date palm trees, their vegetation indices values are presented in Figure 6. Note that "smallish" could be a subjective term as they could be mismanaged, moderately infected, or simply planted later than nearby trees.

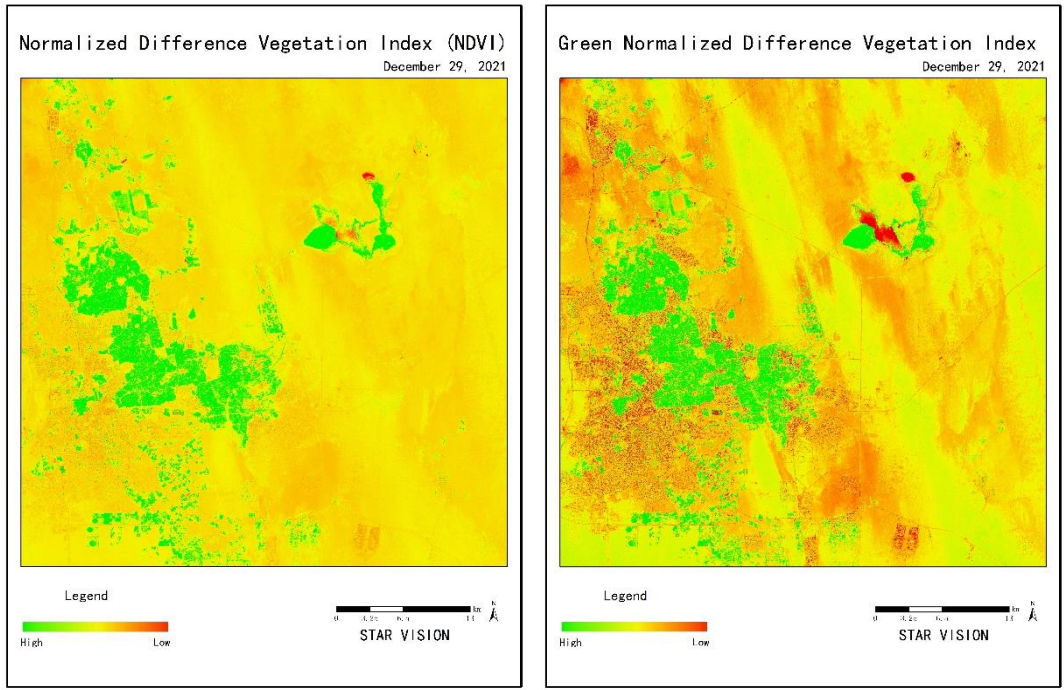

Figure 5 NDVI and gNDVI of study area

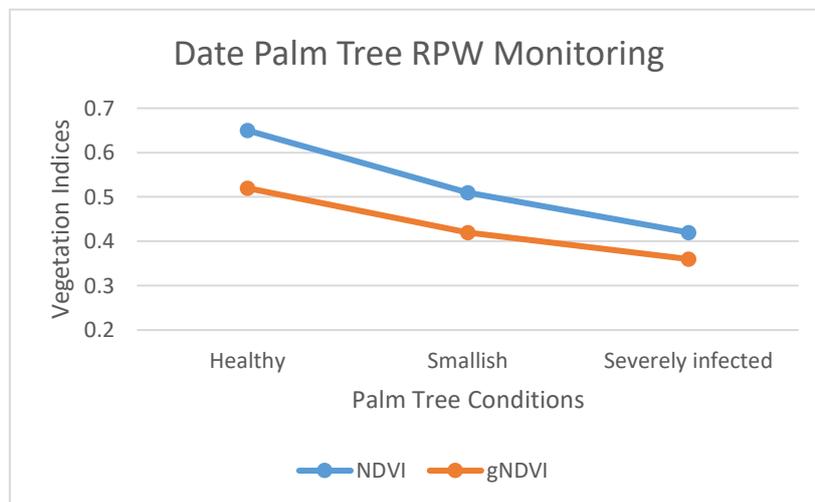

Figure 6 Vegetation indices values across different classes

During the preliminary study by the research team, the magnitude of separation amongst all health status classes was relatively low for vegetation indices. NDVI and gNDVI showed marginally better distinction than other vegetation indices. Thus, by examining these vegetation indices alone cannot suffice for RPW detection at earlier stages. Nonetheless, these results can help locate specific regions where date palms could be largely cultivated. These key values also contributed to classification of palm tree health status.

After selecting specific regions and obtaining VHR satellite imagery, the combination of object detection and semantic segmentation techniques could be utilized to extract each individual palm tree, as shown in Figure 7 below. If a small group of pixels within a window conform to a "high probability" and scored higher than the tree crown detection threshold, each palm tree can be recognized.

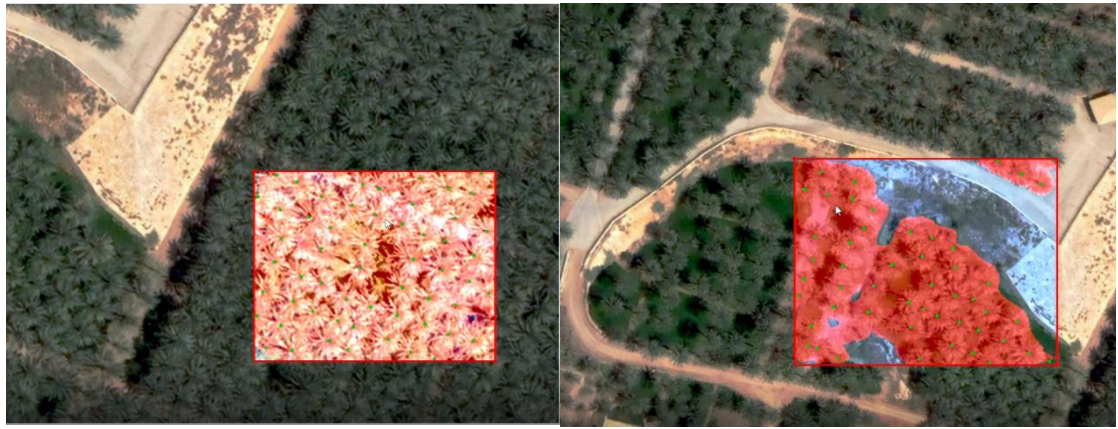

Figure 7 Recognition of individual palm tree

The classification metrics on test data of the proposed method are as follows:

Table 2. Evaluation metrics of proposed method on test data.

| Category | Value |
| --- | --- |
| Number of samples (total) | 4,682 |
| Accuracy | 0.9617 |
| Precision | 0.9551 |
| Recall | 0.9406 |
| F1 score | 0.9471 |

The model yielded promising results, achieving 0.947 F1 score on test data. In fact, the validation scores are very close to training scores, indicating little evidence of overfitting. This could also be the results of using mostly homogeneous data, as the images contained mainly cultivated date palms. Due to time constraints, this trained model was not validated on exogenous data sets, which could further provide efficacy of this approach.

5. Discussions

This research proposed a novel remote sensing approach to recognize and monitor red palm weevil in date palm trees, using a combination of vegetation indices, object detection and semantic segmentation techniques. The study area consists of date palm trees with three classes, including healthy palms, smallish palms and severely infected palms. Even though the healthy class had the highest number of palm trees and lead to an imbalanced problem, the proposed method achieved a 0.9471 F1 score on test data set. This method demonstrates excellent potential for individual date palm tree recognition and multi-class monitoring of growing status using satellite imagery. It could aid leading date palm industries, smallholders and authorities to precisely manage plantation and efficiently deal with infected date palm trees.

However, this model may not be suitable for other types of plant species but could be applicable for other palm trees provided more data is available. Future work will include exploring and developing a more effective generalized date palm tree detection method. Integration of more local information such as treatment method, socioeconomic characteristics could be helpful. More modeling techniques such as feature fusing and refining multi-level feature characteristics and the multiclass-balanced loss module. Furthermore, model validation on more palm tree data sets would be greatly beneficial.

# References

1. Bannari, A., Mohamed, A. M., & El-Battay, A. (2017, July). Water stress detection as an indicator of red palm weevil attack using worldview-3 data. In *2017 IEEE International Geoscience and Remote Sensing Symposium (IGARSS)* (pp. 4000-4003). IEEE.
2. Carter, G. A. (1991). Primary and secondary effects of water content on the spectral reflectance of leaves. *American journal of botany*, *78*(7), 916-924.
3. El-Shafie, H. A. ., Faleiro, J. R., & Aleid, S. M. (2013). Full Length Research Paper A meridic diet for laboratory rearing of Red Palm Weevil , Rhynchophorus ferrugineus ( Coleoptera : Curculionidae ), 8(39): 19241932.
4. European and Mediterranean Plant Protection Organization. (2021.) EPPO datasheets on pests recommended for regulation. Retrieved from https://gd.eppo.int
5. Faghih, A. A. (1996). The biology of red palm weevil, Rhynchophorus ferrugineus Oliv. (Coleoptera, Curculionidae) in Saravan region (Sistan & Balouchistan province, Iran). Applied Entomology and Phytopathology, 63: 16-18.
6. Food and Agriculture Organization of the United (FAO) & International Center for Advanced Mediterranean Agronomic Studies (CIHEAM). (2017). The Scientific Consultation and High-Level Meeting on Red Palm Weevil Management. Retrieved from: http://www.fao.org/3/a-bu018e.pdf
7. Gerson, Uri, Applebaum, Shalom. (2020). Rhynchophorus ferrugineus (Olivier). Retrieved from http://www.agri.huji.ac.il/mepests/pest/Rhynchophorus_ferrugineus/
8. Gitelson, A. A., Merzlyak, M. N., & Lichtenthaler, H. K. (1996). Detection of red edge position and chlorophyll content by reflectance measurements near 700 nm. *Journal of plant physiology*, *148*(3-4), 501-508.
9. Ju, R. T., Wang, F., Wan, F. H., & Li, B. (2011). Effect of host plants on development and reproduction of Rhynchophorus ferrugineus (Olivier) (Coleoptera: Curculionidae). Journal of Pest Science, 84(1): 3339.
10. Kagan, D., Alpert, G. F., & Fire, M. (2021). Automatic large scale detection of red palm weevil infestation using street view images. *ISPRS Journal of Photogrammetry and Remote Sensing*, *182*, 122-133.
11. Kalshoven, L. G. E., Laan, P. A. van der, & Rothschild, G. H. L. (1981). Pests of crops in Indonesia. Van Hoeve, Jakarta : P. T. Ichtiar Baru.
12. Koubaa, A., Aldawood, A., Saeed, B., Hadid, A., Ahmed, M., Saad, A., ... & Alkanhal, M. (2020). Smart Palm: An IoT framework for red palm weevil early detection. *Agronomy*, *10*(7), 987.
13. Kurdi, H., Al-Aldawsari, A., Al-Turaiki, I., & Aldawood, A. S. (2021). Early detection of red palm weevil, rhynchophorus ferrugineus (olivier), infestation using data mining. *Plants*, *10*(1), 95.
14. Marzukhi, F., Said, M. A. M., & Ahmad, A. A. (2020). Coconut Tree Stress Detection as an Indicator of Red Palm Weevil (RPW) Attack Using Sentinel Data. *International Journal of Built Environment and Sustainability*, *7*(3), 1-9.
15. Salama, H. S., Zaki, F. N., & Abdel-Razek, A. S. (2009). Ecological and biological studies on the red palm weevil Rhynchophorus ferrugineus (Olivier). Archives of Phytopathology and Plant Protection, 42(4): 392–399.
16. Yones, M. S., Aboelghar, M. A., El-Shirbeny, M. A., Khdry, G. A., Ali, A. M., & Saleh, N. S.

(2014). Hyperspectral indices for assessing damage by the red palm weevil Rhynchophorus ferrugineus (coleoptera: curculionidae) in date palms. *International Journal of Geosciences and Geomatics*, *2*(2), 2052-5591.

**Appendix.**

Late symptoms of RPW

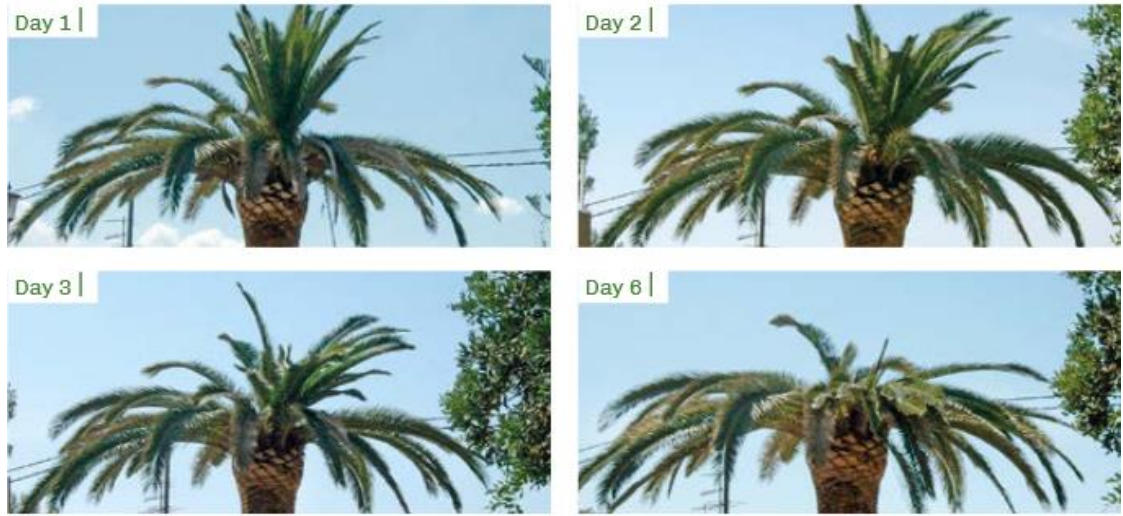

Late symptoms of RPW infestation: although the initial infestation may have started a minimum of three months before, early symptoms may remain undetected for the untrained observer; the final collapse of the palm can take as little as one week (FAO, 2020).

Split training samples

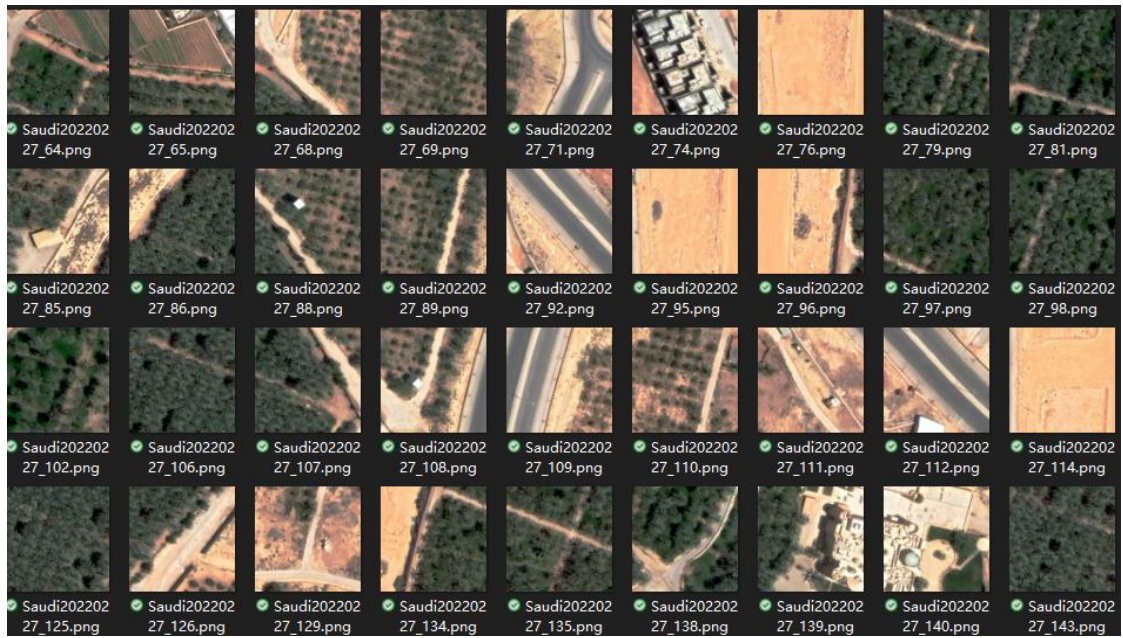

Semantic segmentation examples

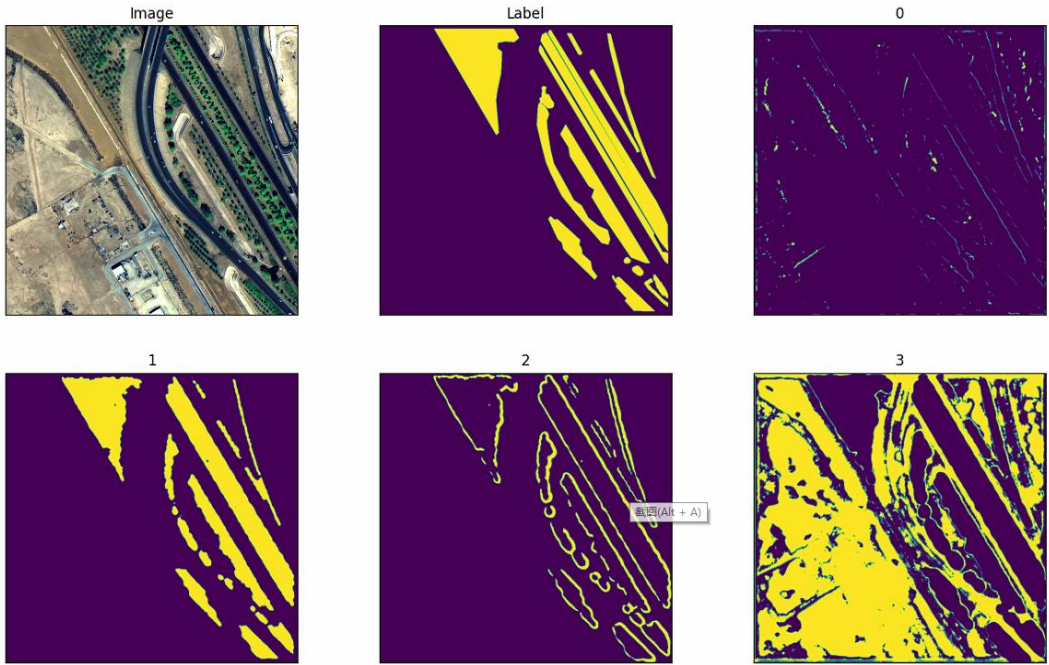

Using inner and outer boundaries, the extracted vegetation area is shown in the bottom left picture.